# IH-ViT: Vision Transformer-based Integrated Circuit Appearance Defect Detection

Xiaoibin Wang [1], Shuang Gao [1], Yuntao Zou[2,*] , Jianlan Guo [1] and Chu Wang [2]

1  Colledge of electronic information of Dongguan Polytechnic, Dongguan 523808, guangdong, China;
2  Huazhong University of Science and Technology, Wuhan, Hubei, China
*  Correspondence: zouyuntao@hust.edu.cn;

**Abstract:** For the problems of low recognition rate and slow recognition speed of traditional detection methods in IC appearance defect detection, we propose an IC appearance defect detection algorithm IH-ViT. Our proposed model takes advantage of the respective strengths of CNN and ViT to acquire image features from both local and global aspects, and finally fuses the two features for decision making to determine the class of defects, thus obtaining better accuracy of IC defect recognition. To address the problem that IC appearance defects are mainly reflected in the differences in details, which are difficult to identify by traditional algorithms, we improved the traditional ViT by performing an additional convolution operation inside the batch. For the problem of information imbalance of samples due to diverse sources of data sets, we adopt a dual-channel image segmentation technique to further improve the accuracy of IC appearance defects. Finally, after testing, our proposed hybrid IH-ViT model achieved 72.51% accuracy, which is 2.8% and 6.06% higher than ResNet50 and ViT models alone. The proposed algorithm can quickly and accurately detect the defect status of IC appearance and effectively improve the productivity of IC packaging and testing companies.

## 1. Introduction

In recent years, the global semiconductor industry has begun to recover, driven by strong demand from emerging applications, mainly 5G communications, automotive electronics, big data, new energy, medical electronics and security electronics. In recent years, although the new crown pneumonia epidemic on the global economy and Integrated Circuit (IC) industry caused adverse effects, but the IC industry is still in the overall growth trend, as shown in **Figure 1**.

The IC industry chain includes several processes including IC design, wafer fabrication, IC packaging and testing, PCB integration, and final product manufacturing, as shown in **Figure 2**. Among them, IC design is including the specific logic and circuit design techniques needed to design integrated circuits. Wafer fabrication is the process consisting of many repetitive sequential processes in semiconductor device manufacturing to produce a complete electronic or photonic circuit on a semiconductor wafer. IC packaging testing is the final stage of semiconductor device manufacturing. packaging is the encapsulation of blocks of semiconductor material in a support housing to prevent physical damage and corrosion. IC testing is the verification of the structure and electrical function of the semiconductor components that have been manufactured to ensure that the semiconductor components meet the requirements of the system.

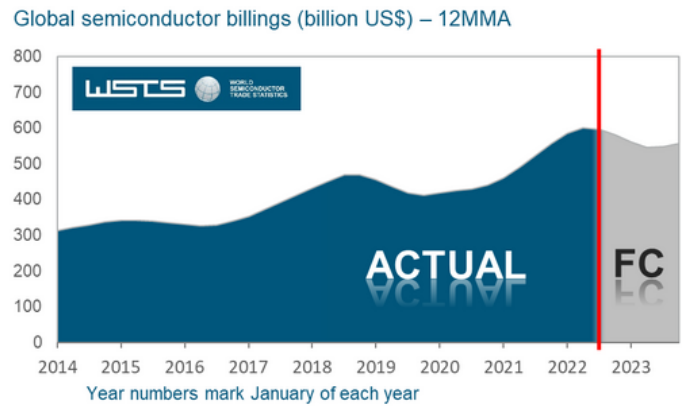

**Figure 1.** Global Semiconductor Market Forecast. (Data Sources: https://www.wsts.org/65/WSTS)

IC testing can be used to pick out products that have functional and cosmetic defects generated during the manufacturing process and provide quality assurance for the products formed subsequently.

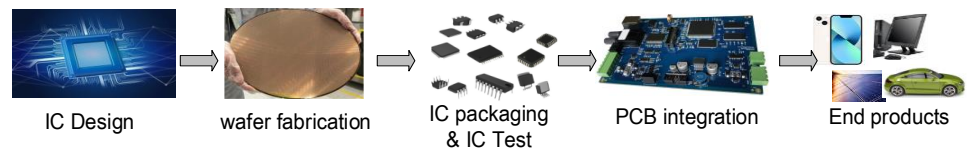

IC Design    wafer fabrication    IC packaging & IC Test    PCB integration    End products

**Figure 2.** IC Manufacturing Process.

In the industrial production process, surface defect detection [1] is a very important part of detecting defective IC appearance in order to ensure the qualified rate and reliable quality of ICs. The traditional method of classifying IC appearance defects is mainly manual inspection [2], but manual inspection relies heavily on the experience of the inspector and cannot objectively and rigorously judge a large number of IC defects, which can neither guarantee the accuracy of the inspection, nor is it efficient enough to meet the requirements of mass production quality inspection. And because the machine vision method [3] in the defect detection has a fast detection speed, high accuracy, flexible application object, strong anti-interference ability and other characteristic, gradually replaced the manual detection. However, at present, in IC appearance inspection, the following difficulties mainly exist.

1. Many types of defects are not easy to identify: In actual industrial production, there are many types of IC appearance defects, such as Surface Scratches, Cover Scratches, Pin Defect, Missing Characters, Unclear Printing, Glue Overflow, etc. As shown in **Figure 3**(a). Therefore, this requires the model to be able to identify a variety of defects. In addition, some new but uncommon appearance defects sometimes appear in the process, which requires the model to have some generalization ability and be able to learn new defects quickly to ensure production efficiency.

2. Process requirements for high detection accuracy, detection accuracy to be improved: With the development of integrated circuits, the size of ICs is getting smaller and smaller, resulting in IC appearance defects more difficult to identify, but the process for certain defects (such as missing characters / character width is not uniform) detection accuracy requirements are very high, for example, in **Figure 3**(b) there is a character width is not uniform defects, due to the small size of the IC, the difference in character width is also very small, the general engineering requirements of the character width defects detection accuracy in the tens of microns level, so this puts forward higher requirements for the detection algorithm.

3. The large variation in the dataset poses difficulties for model learning: the dataset is very difficult to collect due to the very strict control of IC appearance defective products in manufacturing companies, etc. In order to obtain a sufficient dataset, we need to collect images from different sources. These images come from different factories, different data collection equipment, different shooting environments, etc., resulting in non-uniform resolution, size and size, and information density of the collected images, as shown in **Figure 3**(c) and (d). This non-uniformity causes the actual dataset to be very different from the standard dataset (e.g., Image Net), which cannot be directly handed over to the machine learning model for processing. This is also a problem that needs to be solved.

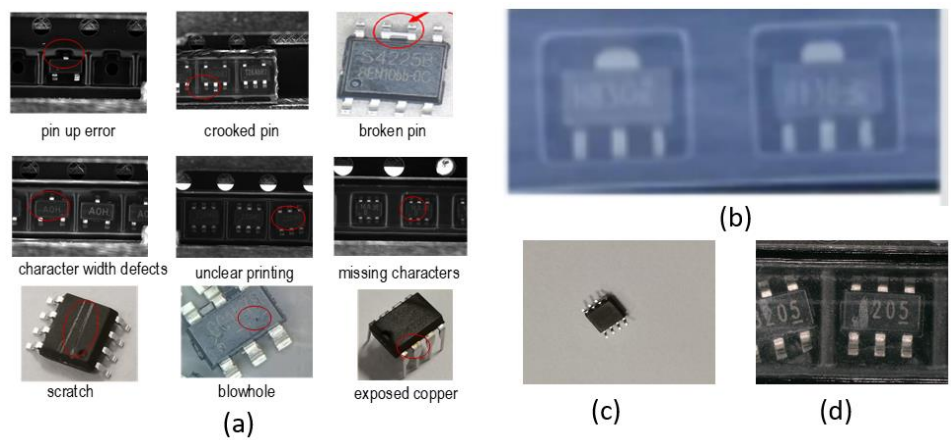

**Figure 3.** IC surface defect sample

In summary, at present, in the IC appearance defect test, there are many types of defects, high detection accuracy requirements, data set collection difficulties and other problems, resulting in the IC appearance defect detection accuracy is not high, detection speed to be improved. In recent years, artificial intelligence, especially deep learning techniques, have achieved good results in image classification and detection tasks. The current deep learning based techniques fall into two main categories: One class is based on CNN [4] convolutional neural network (CNN, Convolutional Neural Network) and the other class is based on Transformer [5]. Convolutional Neural Network (CNN, Convolutional Neural Network) based models have two advantages [6]: one is the sparse connectivity provided by the convolutional layers, as it requires only a small number of weights and connections, saving memory; the other advantage is weight sharing, a feature that helps to reduce the time and cost required for training. Therefore, CNNs excel in terms of latency accuracy on lightweight datasets [7]. However, CNNs have some weaknesses, including slow operation due to the max-pool operation, while not considering the multiple perspectives that can be obtained through learning [8], which leads to ignoring global knowledge. The surprising results of Transformer models on natural language tasks have attracted their application to computer vision problems. One of the significant advantages of Transformer is the ability to model long-term dependencies between input sequence elements and to support parallel processing of sequences [6]. Unlike convolutional networks, Transformer is designed to require minimal inductive bias and is naturally suited as an ensemble function. In addition, Transformers' straightforward design allows multiple modalities (e.g., images, video, text, and speech) to be processed using similar processing blocks and demonstrates excellent scalability for very large capacity networks and huge datasets. These advantages have led to exciting advances in many vision tasks using Transformer networks. Nevertheless, Dosovitskiy A et al. in their study [9] found that Transformer has a significantly lower level of image-specific induction bias compared to CNNs. To overcome the induction bias, the model needs to be trained on a very large and sufficient dataset so that it can identify these image-specific features on its own based on the training examples. The end result is a longer training time, a significantly higher demand on computer resources, and a large number of datasets to process.

The main contributions of this paper are as follows:

1.  Better recognition

The whole anomaly recognition framework well blends the features of convolutional neural network and Transformer. The features are obtained from both local and global aspects to obtain better recognition efficiency. The multi-channel Transformer is also proposed for the problem of uneven information density of images, and different segmentation sizes are used to ensure that high information density regions are effectively segmented, thus improving the recognition efficiency.

2.  Adapting to more complex data sources

Defect detection also has a common problem, the sample imbalance problem. Positive samples are very easy to collect, while abnormal samples are difficult to collect. In order to collect more negative samples, it is necessary to obtain pictures from different companies and different production lines. These pictures have differences in size, resolution, etc. Also more existing images are needed for image enhancement. The image transformation even causes the diversity of image sizes. This model standardizes the format of images from multiple sources and multi-channel segmented images by matrix transformation. It enables a Transformer to handle multi-source, multi-size and multi-channel images.

3.  Stronger generalization capabilities

The IC anomaly detection equipment that is put into use will be placed in various IC production plants around the world. A large number of deployments will certainly keep finding new anomalies. If the model needs to be retrained every time it encounters a new anomaly and then deployed to devices around the world, it would be an extremely complex or even impossible operation and maintenance task. In contrast, the IH-vit model proposed in this paper has a strong generalization capability. New knowledge can be learned by fine-tuning the model, and only a small amount of network and computing power is needed to operate and maintain the global devices. The anomaly detection capability of IC detection devices is continuously improved.

4.  A more compact model

In the engineering field, companies are different from research institutes, who tend to pursue cost effectiveness. Scientific research units design models mainly to pursue better results. While enterprises tend to pursue the best cost performance. So the hardware configuration cannot be used indefinitely like scientific research units. Equipment that is too expensive is unable to get mass sales. So model compression is an important need in industry. The compressed models can run on hardware with lower configuration conditions, thus reducing costs. The model proposed in this paper has been compressed on a large scale in terms of data size and computing power requirements and obtained excellent results. The hardware requirements are reduced while increasing the recognition efficiency.

The article is organized as follows: the second introduces the work related to appearance defect detection and deep learning models applied to appearance detection, the third chapter presents the feature engineering of the dataset, the fourth section presents our model, the fifth chapter presents the experimental results, the sixth chapter presents the conclusion and analysis, and the last chapter presents the contributions of this study.

## 2. Related Work

### 2.1. Appearance defect detection

Surface defect detection refers to the detection of defects such as scratches, flaws, foreign body shielding, and holes on the surface of the sample to be tested, so as to obtain relevant information about the surface defects, such as category, contour, location, and size [10]. Traditional feature-based machine vision algorithms for surface defect detection of industrial products are mainly: in 1995, Bennett, Marylyn Hoy et al. proposed a defect detection method [11], which minimizes the number of defect candidates using the integration of multiple comparison detection results and discriminates based on the normal block image model (DNPM) to determine whether a candidate is defective or normal. This new technical method can detect whether the package IC of BGA has co-planar pins or not, and the IC size can be accurately calculated and the error can be effectively reduced using this method; in 2010, Liu H, Zhou W et al. proposed a two-dimensional wavelet transform feature extraction algorithm for wafer defect detection after studying the difference between the inhomogeneous deposition area on the wafer surface and the background to detect and calibrate it [12]. In 2011, D Gnieser , Tutsch R et al. used the optical detection principle to identify and detect the defects and cracks between the adjacent solder joints of a BGA package IC [13]; In 2011, Lu et al. combined the improved median filter with the Fourier transform to effectively identifies the defects of solder joints [14].In 2012, Kaitwanidvilai S et al. proposed an algorithm for detecting and calculating the number of pins on a IC, they calculated the number of pins by measuring the IC pin characteristics based on the wavelet function transform [15]; In 2014, Berges et al. processed the surface images of MOSFET ICs, and in order to detect the amount of pinholes on the IC package surface, they used the different contrast between the pinhole region and the metal region of the IC image as the object of study, and effectively measured the pinhole number value [16]; In 2020, Yuexin Wen et al. used the improved defect edge scanning method as the core to detect the scratch defects on the bare PCB with copper plating surface, and the accuracy reached 0.1mm, but the modified algorithm could not help for the finer scratches [17]. The algorithm does not work for smaller scratches. The disadvantage of traditional feature-based machine learning methods is that they cannot directly generalize the image data itself to obtain a common feature expression, so there is a need to design a more direct and effective method for extracting common feature expressions from images to face such problems.

The rapid development of deep learning makes it more and more widely used in the field of defect detection. Ding, R et al proposed a multi-layer deep feature fusion method to calculate the similarity between templates and defective circuit boards, the model has a better performance than traditional similarity measures in detecting and locating unknown defects in PCB images Performance [18]. Luan, C. et al. proposed a two-layer neural network for cross-category defect detection without retraining. The network uses (A pairwise Siamese neural network) pairwise Siamese neural network for defect detection. It Learn differences from image-pairs containing specific structural similarities rather than from single images [19]. Anvar, A. et al. proposed a ShuffleDefectNet defect detection system based on deep learning for the detection of steel strip surface defects. The average accuracy of the system on the NEU dataset is 99.75% [20]. Hu, B. et al. proposed an image detection method for PCB defect detection based on deep learning. This method constructs a new network based on Faster RCNN as the backbone of feature extraction to better detect small PCB defects; utilizes GARPN to predict more accurate anchor points and incorporates the residual unit of ShuffleNetV2 [21].

### 2.2. Vision Transformer（ViT）

Transformer mainly uses the self-attentive mechanism to extract intrinsic features [5]. It was first applied to natural language processing (NLP) tasks with good results. A. Vaswani et al. [5] first proposed a converter based on the attention mechanism for machine translation and English constituency parsing tasks. j. Devlin et al. [22] proposed Bert based on Transformer, which obtained state-of-the-art performance on 11 NLP tasks. t. B. Brown et al. [23] et al. proposed GPT model based on Transformer. It achieves robust performance on different types of downstream natural language tasks without any fine-tuning. These Transformer-based models have made a major breakthrough in NLP with their powerful representation capabilities. Due to the powerful representation capability of Transformer, there are also many studies applying it to computer vision,e.g., object detection [24], segmentation [25], video understanding [26], etc., and achieving the best performance on different image recognition benchmarks.

Vision Transformer (ViT) is a Transformer-based architecture proposed for image classification tasks, which directly applies a pure Transformer to sequences of image patches to classify images [27]. ViT uses a global self-attentive mechanism to model the contextual relationships between feature sequences, allowing the network to learn the relationships between feature sequences from a global perspective with powerful feature extraction capabilities. Recently, researchers have started to focus on improving the modeling capability of local information. For example, Transformer-in-Transformer (TNT), Swin Transformer, regionViT, etc. TNT [28]. utilizes inner Transformer block to model the relationship between sub-patches and an outer Transformer block for patch-level information exchange. Swin Transformer [29] performs local attention within a window and introduces a shifted window partitioning approach for cross-window connections. regionViT [30] generates from images regional tokens and local tokens, where local tokens receive global information through the attention of regional tokens. Nowadays, ViT has been applied to various industries. Hütten, N. et al. described in detail the application of ViT in industrial vision inspection [31].Shang H et al. used it for aero-engine blade surface damage detection [32]. However, in the field of IC appearance inspection, no one has obtained good results yet due to various difficulties.

### 2.3. CNN-based model

In 1962, Hubel and Wiesel [33] first introduced a new concept of "receptive field" by studying the visual cortex of the cat brain, which was an important inspiration for the development of artificial neural networks. In 1980, Fukushima [34] proposed a neurocognitive machine and a weight-sharing convolutional neural layer based on the bioneurological theory of receptive field, which was regarded as the prototype of convolutional neural network. In 1989, LeCun [35] invented the convolutional neural network by combining the backpropagation algorithm and the weight-sharing convolutional neural layer, and successfully applied the convolutional neural network to the handwritten character recognition system of the U.S. Post Office for the first time.In 1998, LeCun [36] proposed the classical network model of convolutional neural network, LeNet-5, and again improved the handwritten character recognition correct rate.

In the field of detection CNNs are often mixed with other models.Gao, X. [37] et al. combined FCN with Faster RCNN and designed a deep learning model based on FCN for tunnel defect detection; the model can accurately and quickly detect defects such as stains, leaks and pipe blockages.Xiao, L. et al [38] developed an Image Pyramidal Convolutional Neural Network (IPCNN) model to detect surface defects in images.IPCNN is an improvement of Mask RCNN model, which combines image pyramid and deep convolutional neural network to extract pyramidal features for defect detection.

### 2.4. Comparison of transformer and CNN based models

In this paper, the proposed model is mixed with VIT, so Resnet50 with the same size and effect is chosen. the presence of residual network can solve the gradient problem, and the increase of the number of layers of the network also makes it express better features and correspondingly better performance of detection or classification, plus the use of 1×1 convolution in residual, which can reduce the number of parameters and also reduce the computational effort to some extent.

## 3. Feature Engineering

In the actual industrial production, due to the improvement of technology and process, the proportion of "normal" sample data in the data set is larger, while the amount of

"defective" or "abnormal" sample data is small. This leads to the problem of data imbalance. When training deep learning models, it is often required to have a balanced number of samples of each category in the sample set. When unbalanced samples are applied to a supervised learning task, the algorithm will focus more on the categories with large amounts of data and underestimate the categories with small amounts of data. This can cause the model to fail to learn features that appear to be rarely anomalous and thus fail to detect such anomalies. Therefore, we need to treat the dataset accordingly.

### 3.1. Dataset

In this study, we used the IC surface defect dataset to train and test our IH-ViT model. Our dataset is first described below: The IC surface defect dataset is a dataset with annotations for training, validation and benchmarking of the defect detection algorithm. A total of 2043 original images are included in the dataset, of which 1501 are positive samples without defects and 542 are negative samples with defects. The positive samples include 7 types of ICs.

**错误!未找到引用源。**, and the types of defective images include surface scratches, abnormal cover film scratches, pin problems, missing characters, etc., in a total of 11 categories, as shown in **错误!书签自引用无效。**.

From the above analysis, it can be seen that the ratio of positive samples to negative samples in our dataset is about 3:1, with a smaller percentage of negative samples. Such an unbalanced data set will have a great impact on the training effect. In order to obtain good anomaly recognition, it is necessary to make the data more balanced. This requires data expansion for the negative samples in the dataset.

### 3.2. Data Augmentation

Due to the small number of IC appearance defect dataset, it cannot meet the requirements of model training. Therefore, we need to process the dataset. Here, we use five kinds of data enhancement for data expansion, including image Flip (horizontal and vertical flip), Rotation, Scale, Crop and Translation. The defective data set was expanded from 542 to 7588 sheets as shown in **Figure 4**. Among them, the Flip operation includes horizontal flip and vertical flip, the rotation operation uses two rotations of 90 degrees and 108 degrees, the scale is used to scale outward, the final image size will be larger than the original image size, and then a part is cut out from the new image with the size equal to the original image. Here we perform two scale operations, crop is a random sampling of a part from the original image, and then resize this part to the size of the original image. Here we perform 4 cropping operations. Finally, the Translate operation includes four operations of panning left, right, up and down.

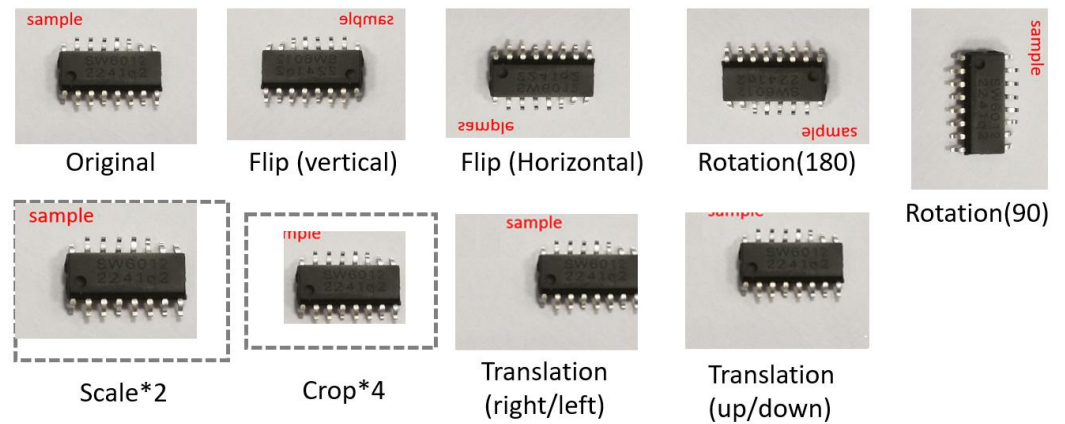

**Figure 4.** Data enhancement method

In this experiment, five data enhancement methods were used for data augmentation of the IC surface defect dataset. In the enhanced dataset, there are 1501 defect-free samples and 7588 defective samples; where the dataset will be divided into training set and test set in the ratio of 8:2, of which 7210 images are used for training and 1879 images are used for testing.

### 3.3. Data Collation

The photos in our dataset, as they are collected through different companies and on different production lines. The different sources of the photos cause differences in size, layout, and many other features. The original dataset includes a variety of image resolutions, the four main ones being 512*480 pixels, 1440*1080 pixels, 4608*3456 pixels and 1276*1702 pixels. In addition, there are a small number of other sizes of photos.

The process of data enhancement is due to the need to perform operations such as cropping of photos. When the relevant operations are performed on the photos in the dataset, more image resolution is generated.

The later transformer framework and CNN framework are designed with strict requirements on the size of the incoming images. In order to meet the data requirements of the subsequent model, we performed the Resize operation on the photos in the dataset after data enhancement. In order to obtain better training results, the photos in the dataset were all resized to 224*224, which is the same size as the photos in the standard ImageNet dataset, and the model can better transfer the previously learned knowledge to the current task.

## 4. Methods

### 4.1. Model Architecture

In this study, we propose an IC appearance defect detection model called IH-ViT (Industry-Hybird-ViT), and the whole model is divided into several parts, such as data enhancement, image resize, ResNet & ViT feature extraction, decision-level fusion, and final classification, as shown in **Figure 5**.

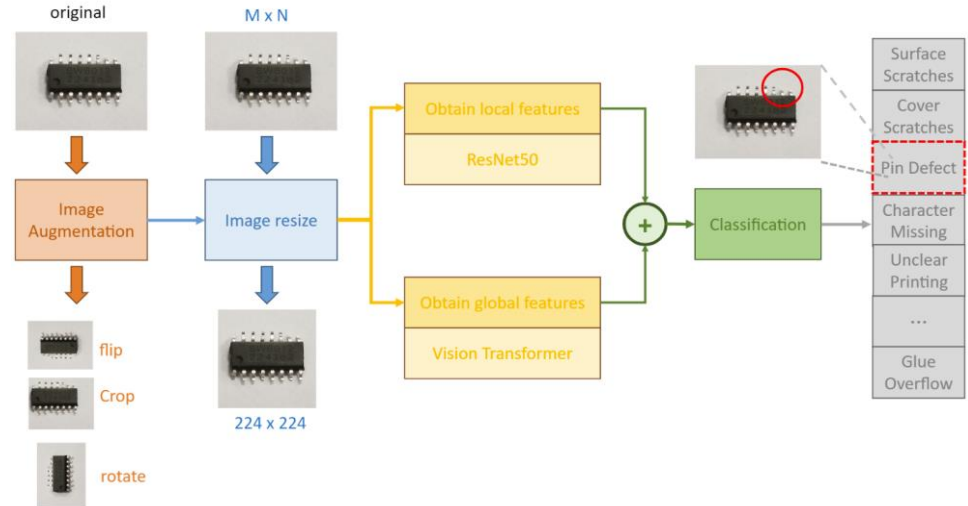

**Figure 5.** IH-ViT: Architecture

### 4.1.1. Data Augmentation

In the IC appearance defect detection task, there is a sample imbalance problem because the ratio of defective samples to defect-free samples is about 1:3 due to the difficulty of obtaining IC appearance defect datasets. The sample imbalance problem in the deep learning model, which leads to the model's classification by a specific preference for a large number of samples, can affect the accuracy of the model. Therefore, firstly, data augmentation techniques are needed to expand the model with data to reduce the adverse effects of the sample imbalance problem on the classification results. Here, we use five data augmentation methods such as flip, crop, and rotate to perform data augmentation on the negative sample dataset.

### 4.1.2. Image Resize

Since the images in the IC appearance defect detection dataset come from different companies, the different shooting conditions lead to great differences in the size, dimensions, and pixels of the images, which brings great inconvenience to the subsequent machine learning models. Therefore, after image enhancement, we have to perform resize operation on the images to convert the different size images into a uniform size of 224*224 pixels to facilitate the subsequent data processing.

### 4.1.3. ResNet & ViT Feature Extraction

After resizing the data, the images are sent to ResNet50 and the improved ViT model respectively for training. The features of IC appearance defective data are extracted separately by taking advantage of the fact that ResNet50 is good at learning local features of images and the improved ViT is good at learning global features.

### 4.1.4. Decision-level Feature Fusion

After ResNet-50 & ViT separate the features extracted from the images, the features are superimposed so that the superimposed features contain both the detailed features of the images and the global features. The final classification is obtained.

### 4.1.5. Final Classification

Based on the superimposed features, the final judgment of the defect category.The main idea of the model is to extract local features and global features by Resnet-50-ViT respectively, and then fuse the feature rows in order to improve the classification accuracy. In addition, the traditional ViT needs to be improved to meet the engineering requirements for the IC appearance defect detection task, where the image information density is uneven, and the IC defects are diverse and difficult to distinguish. We will discuss the ViT improvement in Section 4.2.

### 4.2. Improved ViT structure

A traditional ViT model, as shown in **Figure 6**, cannot handle problems such as uneven image information density and insufficient data collection in industrial production, so we need to improve the ViT model to make sure it meets the project's requirements.

### 4.2.1. Multi-channel segmentation

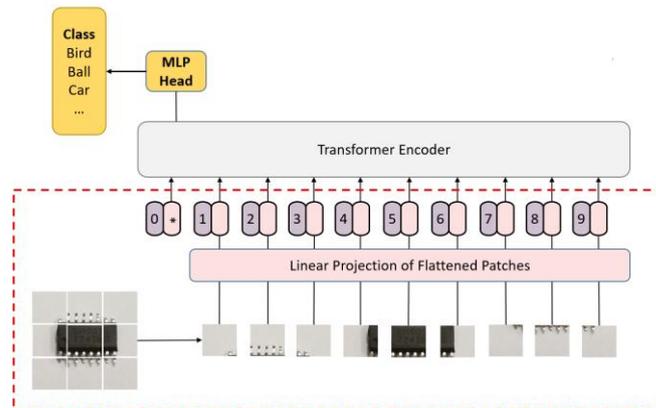

**Figure 6.** Traditional ViT structure

The structure of the traditional ViT is shown in Figure 6, and its core is to cut images into patches according to the P*P matrix, then fuse the segmented patches and position embedding through linear projection, and then directly apply the Transformer model to completely convert image classification into an NLP problem. However, there is a problem in applying the traditional ViT directly to IC appearance defect detection. Since ViT training uses datasets such as ImageNet and Google's Open Images, the image size is more uniform and the image information density is more balanced (i.e., all locations of the image are rich in information).

In our task, on the other hand, since the IC appearance defect dataset is difficult to collect, there are many different sources of images, which makes it difficult to ensure that all images are information-balanced, as in **Figure 7**.(a) where the information is relatively balanced and **Figure 7**.(b) where the information is unbalanced. The problem arises if only a fixed set of P*P matrices are used for cutting. For example, if the same 3*3 matrix is used for segmentation, **Figure 7**.(a) has balanced information density and no problem with the cut, but **Figure 7**.(b) has unbalanced information density, so the cut will make the IC images with more information divided in one patches, while other background images with less information are divided in other patches, which is not conducive to feature learning of the images. Compared with **Figure 7**.(b), **Figure 7**.(c) is a more reasonable way to cut the images.

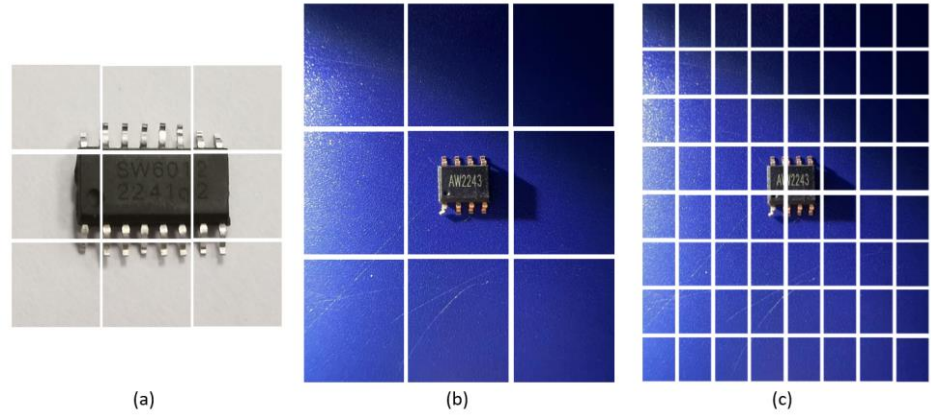

**Figure 7.** Different ways to split images

To solve this problem, we designed multiple segmentation channels, i.e., provide $\{P_1, P_2, \cdots, P_i\}$ multiple segmentation methods to cut the image, to ensure that the regions with high information density can be segmented to different patches. although more segmentation methods can bring more feature learning opportunities but will increase the amount of computation. In the specific process of IC appearance defect detection, companies have to consider the cost performance issue, there is a tradeoff between model accuracy and cost. so we have to find a balance between limited computational power and more cutting methods. Through experiments we found that the two sets of segmentation methods, P=16 and P=32, can be well balanced for our dataset. So we design the model as a two-channel model. If in other scenarios where computational power is not considered, there is also the problem of unbalanced photo information density. It can be designed as a multi-channel model to improve the accuracy.

### 4.2.2. Perform convolution inside each patch

ViT, after segmenting the image, turns the image into a vector by Patch and Position Embedding, and incorporates the position information into the Patches, solving the problem that there is no position information in the self-attention. This can solve the problem of position information between patches, but the position information inside each patch is lost. image information of patches is directly converted from 2D to 1D, and all the position information is lost, as shown in Figure 8, the information between patches 1,2,3...,9 is retained in the figure, but the unknown information inside each patch, such as patch1,or patch 5, the unknown information inside is lost. Although the location information between patches blocks can be compensated to some extent, it still affects the convergence efficiency. In the IC detection task, the photo information density is not balanced, and the loss of internal location information of patches will have a certain impact on the results.

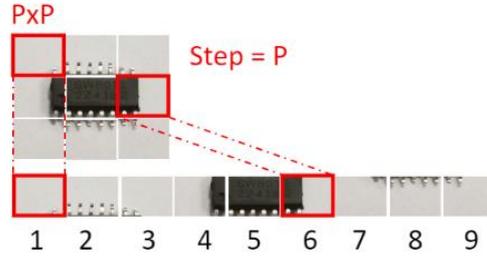

**Figure 8.** patch internal convolution

In fact, when we do the ViT task segmentation, we can also be understood as using a matrix with a field of view of P*P to do a convolution-like operation with a step size of P to perform the segmentation, except that each step does not perform the convolution operation, as shown in **Figure 8**. If we add the convolution operation at this time, i.e., do a convolution with a field of view of P*P and a step length of P, we can obtain the image position information within the patch while segmenting the image. The algorithm is shown in Algorithm 1. It is shown through experiments that such a modification obtains good results for IC detection, which is significantly better than a simple linear transformation.

---

**Algorithm 1 ConvBlock**

input X.size = 16x16
Conv2d(3 , kernel_size = 7, stride = 2, padding = 3)
X.size = 8x8
Relu( )
MaxPool2d(kernel_size = 2, stride = 2, padding = 1)
X.size = 5x5
Flatten( )

---

In addition, we find that by performing convolution inside the patch, we not only obtain the location information inside the patch, but also gain another advantage: after convolution, the size of the patch is compressed while the token remains unchanged.

Taking the original size of 16 x 16 patch as an example.

The original size is $16 \times 16 \times 3 = 768$. After performing the convolution, the size is $5 \times 5 \times 3 = 75$. We use the compression ratio to measure the percentage of the image that is compressed: $compresion\ ratio = \frac{size\ of\ Conv}{Size\ of\ Original} = 9.76\%$, that is, the size of the vector after convolution is only 9.76% of the original, and the compression ratio is about 1:10. Correspondingly, with Token = 196, this reduces the memory by $(16 \times 16 \times 3) \times 196 - (5 \times 5 \times 3) \times 196 = 13528$ , which is a great saving in storage space. It also simplifies the computation process, increases the speed of computation, and reduces the hardware configuration requirements, which is of great significance for engineering applications.

### 4.2.3. Dual channel vector unification

One problem we encounter after changing the ViT model to a two-channel entry is that the size of the vectors changes. The simplest and most common solution is to connect one set of Transformer Encoder behind each one. but in the IC appearance defect detection task, there are constraints on resources such as storage and computation, which makes us need to find a more cost-effective solution. Our solution is to unify the vectors of both channels, so that only one set of transformers can be used to handle the vectors of different channels,as shown in **Figure 9**.

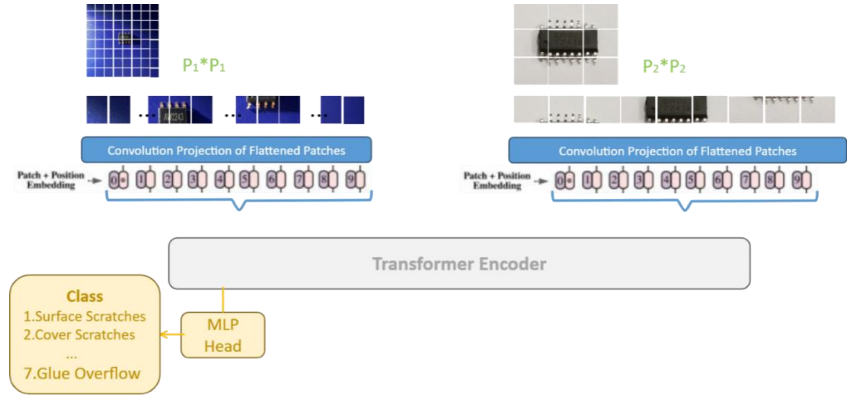

**Figure 9.** Two-channel vector unification

In the image resize process, we process the image to a size of $224 \times 224$ in order to facilitate later processing, i.e., $Fresize(X_{nxm}) = X_{224x224}$.

We set the first partition size $P_1$ to 16 and the second partition size $P_2$ to 32, i.e.,

$$P_1 = 16，P_2 = 32.$$

After the convolution,

$$F_{conv}(P_1.Size.16x16) = P_1.Size.5 \times 5，$$

$$F_{conv}(P_2.Size.32x32) = P_2.Size.16x16.$$

Now, the size of $P_1$ and $P_2$ becomes 75 and 768 respectively, i.e.,

$$Size.P_1 = 5 \; x \; 5 \; x \; 3 \; = \; 75,$$

$$Size.P_2 = 16 \; x \; 16 \; x \; 3 \; = \; 768.$$

The token of the image after partitioning is as follows:

$$No.P_1 = 224 \; x \; 224 \; / \; 16 \; x \; 16 \; = \; 196，$$

$$No.P_2 = 224 \; x \; 224 \; / \; 32 \; x \; 32 \; = \; 49.$$

The token number of the partitioned image Patch1 is 196, and the token number of the partitioned image Patch2 is 49.

$$Patch16 \; (196 \; x \; 75) \; x \; (75x \; 75) \; = \; 196 \; x \; 75$$

For a 16x16 patch, we use a 75*75 matrix for linear changes. In this way, ViT still receives a 196x75 vector.

In order to allow the P2(32x32) patch to use the same Transformer Encoder, that is

$$Patch32 (49 \; x \; 768) x \; (768 x \; 75) \; = \; 49 \; x \; 75$$

We use a (768x75) matrix to convert the size of P2 to 49x75. This can be directly processed by the same Transformer Encoder.

### 4.3. Model Advantages

In summary, we propose an IH-ViT model with Resnet-50 and ViT fusion for the IC appearance defect detection task. The main idea of this model is to extract local features and global features with the advantages of each of Resnet-50 and ViT, respectively, and then fuse the feature rows superimposed to determine the category of defects, which can improve the classification accuracy. In addition, for the IC appearance defect detection task, the image information density is unbalanced, and there are many types of IC defects and it is difficult to distinguish them, the traditional ViT is improved by performing convolution within patch, multi-channel segmentation, etc., which has improved the classification accuracy and compressed the model to reduce the cost to meet the engineering requirements.

The main advantages of our model are as follows:

1. We propose a CNN and Transformer fusion model IH-ViT, which can acquire image features from both local and global aspects, thus better capturing the detailed features of different classes of IC appearance defects and thus obtaining better defect recognition efficiency.

2. We improve the traditional ViT for the problem of uneven information density of images in the IC appearance defect task dataset, and propose a multi-channel transformer. by segmenting the images with different segmentation sizes, we ensure that the high information density regions in the images are effectively segmented, thus further improving the recognition efficiency.

3. By performing the convolution operation inside Patch, it not only makes the model capture the details of IC appearance defects better, but also compresses the model so that it can reduce the cost of computation and operation, which is more suitable for engineering applications.

## 5. Experiment

### 5.1. Experimental setup

We evaluated the performance of the proposed method using three experimental setups. First, we compare the proposed hybrid model with the ResNet50 model alone and with the ViT model alone, with the aim of assessing whether the hybrid model is superior to the model alone. Second, we compare the improved ViT that performs convolution inside the Patch with the standard ViT model, with the aim of assessing whether the improved ViT model that performs convolution inside the Patch has a higher recognition accuracy than the standard ViT model. Third, we compared the improved ViT model that used 2 Patch sizes of cut images with dual channels with the standard ViT that used a single size of Patch cut size, with the purpose of evaluating whether multi-channel cut images can improve the recognition accuracy of the model.

### 5.2. Implementation Details

The learning rate of this method is initialized to 0.001 and decays using the cosine learning rate. We performed 200 epochs of training using the Adam optimizer because the Adam optimizer allows the network to converge faster and adjust the parameters more easily.

Since the model proposed in this study is a hybrid model based on ResNet50 and ViT, the loss function calculation is different from the normal model. The loss functions of the two models need to be calculated

$$Loss = \sum_{i=1}^{i=n} |W_i Loss_{modeli}|/\text{n}$$

$$W1 = a_{ResNet50}$$

$$Loss1 = L_{ResNet50}$$

$$W2 = a_{ViT}$$

$$Loss2 = L_{ViT}$$

$$Loss = |a_{ResNet50} L_{ResNet50} + a_{ViT} L_{ViT}|/2$$

Here, $a_{ResNet50}$ and $a_{ViT}$ are the model weight coefficients, which we can train by linear regression or other methods. In this experiment, for simplicity, the weights of the two models are assumed to be] the same, and $a_{ResNet50} = a_{ViT} = 1$. The weights can be optimized later to improve the effect. The final loss function is obtained: $Loss = |L_{ResNet50} + L_{ViT}|/2$, as shown in 错误!未找到引用源。.

For the ViT model, GELU is used as the activation function along with the L2 regularizer, and a rectified linear unit (ReLu) is used in the CNN along with the L2 regularizer. To prevent bias in the results, the same parameter settings are used for all comparisons. We implemented the method using the Pytorch toolbox, and the momentum coefficient of the network was set to 0.9.

Our experiments are run on a server with the following main configurations: 2 Xeon 10-core processors, 2 3090 graphics cards with 24G RAM (48G total), 4 Samsung 32G RAM, a 500G SSD, and a 4T data disk.

*5.3. Experimental Results*

5.3.1. Comparison of hybrid model and single model

To evaluate the performance of our proposed hybrid model, we compared it with the ResNet50 model alone and with the ViT model alone. Here, the original ViT is used in the hybrid model here instead of the improved ViT in order to prevent other interferences. the comparison metric is acc,which is calculated as follows.

$$acc = \frac{right}{all}$$

The results of the experiments are shown in **Table 1.** Comparison of IH-ViT with ResNet50 and ViT models.

**Table 1.** Comparison of IH-ViT with ResNet50 and ViT models.

| Network | Acc |
|---------|-----|
| ResNet50 | 69.71% |
| ViT | 66.45% |
| IH-ViT | **72.51%** |

Through the above experiments, we have the following findings.

1. ResNet50 obtains 69.71% accuracy, while ViT only obtains 66.45% accuracy, which is a 3.26% improvement. This illustrates the advantage of ResNet50 over ViT on small-scale datasets.

2. The hybrid model IH-ViT achieved 72.51% accuracy, which is 2.8% and 6.06% higher than the ResNet50 and ViT models alone, respectively, showing better performance.
   In summary, the hybrid model IH-ViT has higher accuracy than the separate ResNet50 , ViT models.

5.3.2. Comparison of Improved ViT with Standard ViT Performing Convolution

To evaluate the improvement of model performance after performing convolution inside the 16x16 patch, we compared the ViT after performing convolution with the standard ViT model, and the experimental results are shown in **Table 2.** Comparison of ViT for performing convolution with standard ViT..

**Table 2.** Comparison of ViT for performing convolution with standard ViT.

| Network | Acc |
|---------|-----|
| ViT | 66.45% |
| ViT+Conv | **69.18%** |

With the above experiments, we have the following findings: the accuracy of the model after performing convolution inside the 16x16 patch is 69.18, which is 2.73% higher than the standard ViT of 66.45%, and the performance of the model has been improved.

In addition, we also compare these details to local details by transfomer's multi-headed attention on the whole, and it is easier to find pinning errors, a class of inconsistent errors. This is difficult to do with traditional CNNs that only do local analysis.

5.3.3. Multi-channel ViT versus standard ViT

To evaluate how the improved multi-channel ViT affects the model performance, we compared the ViT with two channels (i.e., $16 \times 16$ patches and $32 \times 32$ patches) with the standard ViT, and the experimental results are shown in **Table 3.** Comparison of multi-channel ViT with standard ViT..

**Table 3.** Comparison of multi-channel ViT with standard ViT.

| Network | Acc |
|---------|-----|
| ViT | 66.45% |
| 2channel-ViT | **67.83%** |

From the experimental results, the accuracy rate after using 2channel-ViT to de-cut photos is 67.83, which is 1.38% higher than the standard ViT's 66.45%, and also improves some accuracy.

## 6. Conclusion and Analysis

### 6.1. Conclusion

In this study, we developed a hybrid model based on CNN and ViT for IC appearance detection. The proposed model takes advantage of both CNN and ViT to acquire image features both locally and globally, which results in better accuracy of IC defect recognition. We improve the traditional ViT by performing an extra convolution operation inside the patch in order to distinguish IC appearance defects from details more accurately. We implemented class-wise data balancing based on image enhancement to compensate for the imbalance in the number of IC defect-free and defective samples in the IC appearance dataset. This avoids the problem that the proposed model has a preference for a given class with a large amount of data, which negatively affects the detection results. To address the information imbalance problem of samples due to the diversity of data set sources in the IC appearance defect detection task, we employ a dual-channel image segmentation technique to further improve the IC appearance defect accuracy.

1. Through experiments, we found that the proposed IH-ViT hybrid model achieved 72.51% accuracy, which is 2.8% and 6.06% higher than the ResNet50 and ViT models alone, respectively, indicating that our proposed hybrid model is effective in improving IC appearance defect recognition.

2. Through the implementation, we found that the accuracy of the model after performing convolution inside the 16x16 patch is 69.18, which is 2.73% higher than the standard ViT's 66.45%, and the performance of the model has been improved.

3. From the experimental results, the accuracy after using 2channel-ViT to de-cut the photos is 67.83, which is 1.38% higher than the standard ViT of 66.45%, a significant improvement in accuracy.

This indicates that our proposed model, in improving the recognition rate of IC appearance defects, is effective.

### 6.2 Analysis

1. The hybrid model improves the accuracy of detection more than the separate model, which is in line with expectations, because the hybrid model combines the advantages of ResNet50 local learning capability and ViT global learning capability, and thus can better capture the features of IC appearance defects and obtain higher accuracy.

2. We believe that the accuracy can be improved after performing convolution inside the patch for the following reasons: the task of this study is IC appearance defect detection, and the difference between defective samples, as well as between defective and normal samples is mainly in details, which is not very large and thus still different from ordinary classification tasks. In a normal classification task, such as classifying two animals or two fruits, there are distinguishable features in the whole and in the local area. In this project, the difference defect is mainly in local and detail, so after performing convolution inside the patch, more detail information is obtained, which improves the accuracy.

3. The recognition accuracy is also improved by 1.38% after executing the image segmentation with dual channels. The accuracy improvement is limited because the engineering cannot provide too much computational power, and it is still difficult to handle images with complex sources just in the form of two channels. In addition, the improvement of two channels is only for the case where there is a large amount of unbalanced picture information in this dataset; if the photo quality is high and the information distribution is balanced, it is not necessarily possible to obtain a significant improvement

### 6.3. Shortcomings and Improvements

In this study, we improve the accuracy of IC appearance defect detection by a hybrid model and the improvement of ViT. However, in fact, there are still the following shortcomings in our work, for example: the number of datasets is not large enough, especially the number of defect datasets is still small; the sample quality of datasets is not high enough, and there is the problem of unbalanced picture information.

In response to the above problems, in the next step, we will further carry out data collection work, firstly, to continuously improve the quantity and quality so that the model can learn the features better.

Secondly, due to the limitation of computational power, we can only use the two-channel method to segment the images, but it is difficult to get good segmentation effect for various photos. If the segmentation is done in an adaptive way, better results will be achieved. Moreover, it is possible to optimize the dual channels into channels. It can save computational power and get better results.

The third is that the model loss function is only calculated with 50% of each weight. And here if you want to better use the advantages of the two models, you can train the weights. The model can dynamically adjust the weights of both algorithms according to the different inputs. This way, we can get better and reasonable processing weights for different photos and thus get better model results.

## 7. Contribution

In this study, a hybrid model based on IH-ViT is proposed for IC appearance defect detection. The whole model incorporates a good blend of CNN and Transformer features to obtain features from both local and global aspects to obtain better IC defect recognition accuracy. For the problem of diverse data set sources and low image quality (uneven information) in IC appearance defect detection tasks, a multi-channel transformer is proposed to ensure that high information density regions are effectively segmented by different segmentation sizes, thus improving the recognition efficiency. In addition, by improving the traditional ViT, i.e., adding convolutional operations within the patch, it not only makes the accuracy of the model improved, but also compresses the model and reduces the engineering cost. Through experimental evaluation, our proposed model improves the accuracy by 3.26%.

Although the accuracy of the proposed model defect detection in this paper is only around 70%, it can still recover a significant economic loss compared to manual sampling detection. The calculation is as follows.

Assumptions:

1. The total annual IC production value is I ICs; I total production = 1.15 trillion
   - The annual percentage of abnormal ICs p%; traditional IC abnormal rate of 0.3% to 1%, we take p% = 0.5% calculation

2. Defect detection accuracy rate is acc%; the latest readiness rate of this thesis is acc% = 73.51%

3. Economic loss per defective IC is L; we just calculate the logistics and rework cost L = 1 USD
   Abnormal IC detection volume H defective rate

= (annual IC yield $\times$ IC defect rate) $\times$ detected accuracy rate

= (I total production $\times$ p% ) $\times$ acc% = 4.2 billion ICs

Economic loss recovered

= (annual IC production $\times$ IC defect rate) $\times$ detected accuracy $\times$ economic loss per IC

= (I total production $\times$ p% ) $\times$ acc% $\times$ L defect rate

= \$4.2 billion

This paper differs from traditional research in that it is not simply pursuing accuracy rate. Rather, we are pursuing the cost/performance ratio between the production price, usage cost and failure accuracy of the IC inspection equipment. If we simply pursue the failure accuracy rate, we have a better solution. But considering the industrial production process, more and more unknown errors will be found. The speed of detection also needs to reach 100 pieces/second, and we had to make some unique designs in the architecture design. And these designs arrived at the design expectations of the project.